\documentclass[table]{article} 
	\usepackage{iclr2018_conference,times}
	\usepackage{hyperref}
	\usepackage{url}
	
	\usepackage{xcolor}
	\usepackage{booktabs} 
	\usepackage{lineno,hyperref}
	\usepackage{epsfig,graphicx}
	\usepackage{subcaption}
	
	\usepackage{epstopdf}
	
	\usepackage{amsmath,amscd,amsbsy,amssymb}
	\usepackage{latexsym,mathrsfs,url}
	\usepackage{enumitem,balance,mathtools}
	\usepackage{wrapfig}
	\usepackage{mathrsfs, euscript}
	\usepackage{multicol}
	\usepackage{float}
	\usepackage{enumitem}
	\usepackage{amsthm}
	\usepackage[noend]{algpseudocode}
	\usepackage{algorithm}
	\usepackage{lscape}
	\usepackage{multirow}

  \title{The Local Dimension of Deep Manifold}
	
\small{	
\author{
	Mengxiao Zhang\thanks{The first four authors contributing equally.}  ~\& Wangquan Wu \& Yanren Zhang \& Kun He\thanks{Corresponding author.} \\
	Department of Computer Science \\
	Huazhong University of Science and Technology \\
	Wuhan 430074, China \\
	\texttt{\{zmx, u201514497, hhxjzyr, brooklet60\}@hust.edu.cn} \\
	\AND
	Tao Yu, Huan Long \\
	Zhi Yuan College / Department of Computer Science and Engineering, Shanghai Jiaotong University \\ 
	\texttt{ydtydr@sjtu.edu.cn, longhuan@cs.sjtu.edu.cn} \\
	\AND
	John E. Hopcroft \\
	Department of Computer Science\\
	Cornell University, \\
	Ithaca, NY 14853, USA \\
	\texttt{jeh@cs.cornell.edu}
}

\iclrfinalcopy 

\begin{document}

	\maketitle
	
	\begin{abstract}
	Based on our observation that there exists a dramatic drop for the singular values of the fully connected layers or a single feature map of the convolutional layer,
	and that the dimension of the concatenated feature vector almost equals the summation of the dimension on each feature map,
	we propose a singular value decomposition (SVD) based approach to estimate the dimension of the deep manifolds for a typical convolutional neural network VGG19. 
	We choose three categories from the ImageNet, namely Persian Cat, Container Ship and Volcano, 
	and determine the local dimension of the deep manifolds of the deep layers through the tangent space of a target image.
	Through several augmentation methods, we found that the Gaussian noise method is closer to the intrinsic dimension,
	as by adding random noise to an image we are moving in an arbitrary dimension, and when the rank of the feature matrix of the augmented images does not increase
	we are very close to the local dimension of the manifold. 
	We also estimate the dimension of the deep manifold based on the tangent space for each of the maxpooling layers. 
	Our results show that the dimensions of different categories are close to each other and decline quickly along the convolutional layers and fully connected layers.
	Furthermore, we show that the dimensions decline quickly inside the Conv5 layer. 
	Our work provides new insights for the intrinsic structure of deep neural networks and helps unveiling the inner organization of the black box of deep neural networks.
	\end{abstract}
	
	\section{Introduction}
	\label{secintroduction}
	To have a better understanding of deep neural networks, a recent important trend is to analyze
	the structure of the high-dimensional feature space. Capitalizing on the 
	\textit{manifold hypothesis} (\citealt{cayton2005algorithms}; \citealt{narayanan2010sample}), the distribution of 
	the generated data is assumed to concentrate in
	regions of low dimensionality. In other words, it is assumed that
	activation vectors of deep neural networks lie on different low
	dimensional manifolds embedded in high dimensional feature
	space. 

	Note that the rationality of many manifold learning algorithms
	based on deep learning and auto-encoders is that one learns an
	\textit{explicit or implicit coordinate system} for leading
	factors of variation. These factors can be thought of as concepts
	or abstractions that help us understand the rich variability
	in the data, which can explain most of the structure in the
	unknown data distribution. See
	\citet{Goodfellow-et-al-2016} for more information.

	The dimension estimation is crucial in determining the number
	of variables in a linear system, or in determining the number of
	degrees of freedom of a dynamic system, which may be embedded
	in the hidden layers of neural networks. Moreover, many algorithms
	in manifold learning require the intrinsic dimensionality of the
	data as a crucial parameter. Therefore, the problem of estimating
	the intrinsic dimensionality of a manifold is of great importance, 
	and it is also a crucial start for manifold learning.

	Unfortunately, the manifold of interest in AI (especially for deep
	neural networks), is such a rugged manifold 
	with a great number of twists, ups and downs with strong curvature. 
	Thus, there is a fundamental difficulty for the manifold
	learning, as raised in ~\citet{bengio2005non}, that is, if the manifolds are
	not very smooth, one may need a considerable number of training examples 
	to cover each one of these variations, and there is no chance 
	for us to generalize to unseen variations. 

\begin{figure}[htb]
	\centering
	\includegraphics[width=0.45\textwidth]{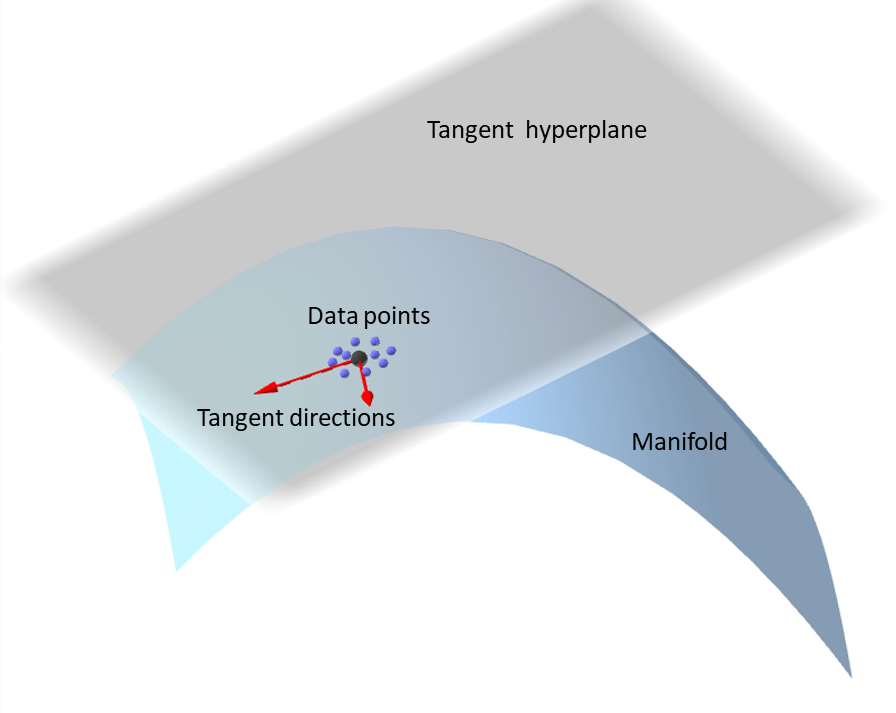}
	\vspace{-1.5em}
	\caption{A two-dimensional manifold with a small region where
		data points concentrate, along with a tangent plane and associated
		tangent directions, forming a basis that specifies the
		directions of small moves one can make to stay on the
		manifold.}
	\vspace{-1em}
	\label{figtangentspace}
	\end{figure}

	Our work is based on an important characterization of the manifold, namely,
	the set of its \textit{tangent hyperplanes}. For a point $p$ on a
	$d$-dimensional manifold, the tangent hyperplane is given by a local basis 
	of $d$ vectors that span the local directions of variations allowed
	on the manifold. As illustrated in Figure \ref{figtangentspace}, these
	local directions specify how one can change $p$ infinitesmally
	while staying on the manifold.

	Based on above analysis, our work focuses on a thorough
	exploration of the local hyperplane dimension of the activation manifold 
	in deep neural networks. 
	Creating an artificial data cluster concentrated in
	regions of the local tangent hyperplane, we apply SVD to
	the data cluster in different layers or feature maps in neural networks. 
	Through thorough analysis, we reach the following fascinating results.

	\begin{itemize}
	\item There exists a dramatic drop for the singular values of the fully connected layers or a single feature map of the convolutional layer.
	\item For convolutional layers, the dimension of the concatenated feature vector almost equals the summation of the dimension on each feature map.
	\item The dimensions of different image categories are close and the dimension declines quickly along the layers. 
	\end{itemize}

	To our knowledge this is the first thorough exploration of 
	manifold dimension on very deep neural networks. We wish our
	work sheds light on new understandings and inspires further investigations
	on the structure of manifolds in deep neural networks. 

	\section{Related Work}
	\label{secrelatedWork}

	With the great success of deep learning in many applications including computer vision and machine learning,
	a comprehensive understanding of the essence of deep neural networks is still far from satisfactory. 
	Related works can be classified mainly into three types. The first kind of work focuses on the difference of random networks and trained networks~\citep{SaxeICML11, he2016powerful,rahimi2009weighted}.
	There are also works that focus on the theoretical understanding of learning~\citep{zhang2017ICLR, LiYuanNIPS2017}, while the rest focus on the inner organization or feature representations through visualization~\citep{Mahendran2015CVPR,Dosovitskiy2015inverting}. 

	Up until now, there are only a few works in exploring the property of deep manifolds formed by activation vectors of the deep layers. In this section, we highlight the most related work for manifold learning and dimension determination.

	Manifold learning has been mainly applied in unsupervised learning procedures that attempt to capture the manifolds~\citep{van2008visualizing}. It associates each of the activation nodes with a tangent plane that spans the directions of variations
	associated with different vectors between the target example and its neighbors. \citet{weinberger2004learning} investigate how to
	learn a kernel matrix for high dimensional data that lies on or
	near a low dimensional manifold. \citet{rifai2011manifold}
	exploit a novel approach for capturing the manifold structure
	(high-order contractive auto-encoders) and show how it builds a
	topological atlas of charts, with each chart characterized by
	the \textit{principal singular vectors} of the Jacobian of a
	representation mapping. \citet{kingma2014semi} propose a
	two-dimensional representation space, a Euclidean
	coordinate system for Frey faces and MNIST digits, learned by
	a variational auto-encoder.

	There are several efficient algorithms to determine 
	the intrinsic dimension of high-dimensional data.
	Singular Value Decomposition (SVD), also known as Principal Component
	Analysis (PCA), has been discussed thoroughly in the literature~\citet{strang1993introduction}. 
	In applications, the choice of algorithm will rely on the geometric prior of the given data and the expectation of the outcome.  
	In addition, researchers have also proposed several improved manifold-learning algorithms considering more pre-knowledge of the dataset.  For example, 
	\citet{5278634} estimate
	the intrinsic dimensionality of samples from noisy low-dimensional
	manifolds in high dimensions with multi-scale SVD.
	\citet{NIPS2004_2577} propose a novel method for estimating the
	intrinsic dimension of a dataset by applying the
	principle of maximum likelihood to the distances between close
	neighbors. \citet{Haro:2008:TPM:1416831.1416842} introduce a
	framework for a regularized and robust estimation of
	non-uniform dimensionality and density for high dimensional noisy
	data.  

	To have a comprehensive understanding of the manifold structure of 
	neural networks, it is natural to treat the activation space 
	as a high-dimensional dataset and then characterize it by determining 
	the dimension. This will give us a new view of the \emph{knowledge} 
	learnt from the neural network and hopefully the information hidden inside. 
	However, to the best of our knowledge, there is nearly no 
	related work in determining the intrinsic dimensionality 
	of the deep manifold embedded in the neural network feature space.  
	In the following we will give a thorough study on this topic.
	
	\section{Manifold Dimension Characterization}
	\label{secdimension}
	In this section we describe our strategy to determine 
	the dimension $d$ of the local tangent hyperplane of the manifold, 
	along with necessary definitions and conventions which 
	will be used in specifying the dimension of the manifold dataset.
	
		\subsection{determine dimension by svd}
		\label{secSVD}
		It is known that if there is a set of data $\{x_{i}\}_{i=1}^{n}$
		that can be regarded as a data point cluster and lie
		\textit{close} to a \textit{noiseless} $d$-dimensional
		hyperplane, then by applying SVD, the number of non-trivial singular
		values will equal to $d$ --- the intrinsic dimension of the data
		point cluster. 
		In the context of a manifold in a deep neural network, 
		a cluster of activation vectors pointing to a manifold 
		embedded in feature space $\mathbb{R}^D$, can be approximated
		as concentrated on a $d$-dimensional tangent hyperplane,
		whose dimension directly associates with the manifold.
		
		However, the challenge here in dimension estimation is
		that noise everywhere are influencing the dataset
		making it hard to get the correct result:
		\begin{enumerate}
		\item \citet{5278634} point out that when D-dimensional noise is
		added to the data, we will observe :$\tilde{x}_i= x_i+\eta_i$,
		where $\eta$ represents noise. The noise will introduce
		perturbation of the covariance matrix of the data, which will
		lead to the wrong result.
		\item \citet{Goodfellow-et-al-2016} also mentioned that some
		factors of variation largely influence every single piece of
		the observed data. Thus those factors we do not care about 
		(or simply considered as noise) may lead researchers to wrong result
		with high probability.
		\end{enumerate}
		
		To solve the above problems, we make use of the following observations:
		\begin{enumerate}
		\item By introducing representations that are expressed in terms
		of different, simpler representations, deep neural networks extract
		high-level, abstract features from the raw data, that make it
		successfully disentangle the factors of variation and discard the
		ones that we do not care about (noise), see
		\citet{Goodfellow-et-al-2016} for more information. 
		It turns out that the noise in feature space will be so small
		that we are likely to have the singular values from factors we care about 
		significantly larger than the remaining singular values generated by noise 
		and get the right result.
		\item \citet{johnstone2001} and many works have shown that when
		$n$, the number of data points, goes to infinity, the behavior
		of this estimator is fairly well understood.
		\end{enumerate}
		
		Based on the above analysis, we propose the following solution: 
		\begin{enumerate}
		
		\item By using a pre-trained deep neural network, after the
		feed-forward process, the feature vectors have little
		irrelevant noise remaining and preserve all useful factors of variation.
		So we can make sure that feature vectors lie on a noiseless
		manifold embedded in high dimensional feature space.
		
		\item By introducing some picture augmentation methods, we will
		generate considerable amount of similar pictures(also classified as the same
		class with high probability by deep neural network), then we
		will get a sufficiently large cluster of feature vectors that lie
		\textit{close} to a local tangent $d$-dimensional
		hyperplane of the noiseless manifold.
		
		\item Finally we apply the original SVD on this feature vector
		cluster which lies \textit{close} to \textit{noiseless}
		local tangent $d$-dimensional hyperplane and give a precise estimation 
		of the local dimension of the manifold.
		
		\end{enumerate}
		
		Following paragraphs give a more formal description of our
		solution to computing the dimension.
		
		Let $X_n=\{x_i\}^n_{i=1}$ be the set of image data points we
		generate, $x_1$ is the original image classified by the neural
		network as a specific class with high probability,e.g
		$P(x_1)>0.99$. By using augmentation methods, we generate $n-1$
		augmented images and keep all augmented images
		classified to be in the same class with high probability.
		$P(x_i)>0.99$, $i=2 \cdots n$.
		
		Let $\lambda=\mu+\eta$ be augmentation information we have introduced
		to the image. The $\lambda$ can be divided into two components:
		some irrelevant noise($\eta$) combined with useful factors of
		variance($\mu$), let $f$ be the underlying network feature
		extract function. After feed-forward process in a specific layer,
		$n$ feature vectors in $\mathbb{R}^{D}$ are denoted as a $D \times
		n$ matrix $A_{D,n}$. For simplicity, we denote $A_{D,n}$ as
		$A_n=\{a_i\}^n_{i=1}$. $\mathcal{P}$ is the local approximate
		hyperplane of manifold $\mathcal{M}$.
		
		In the real image space, we have $\tilde{X}_{n}=\{
		x_{i}+\sigma_{i}\}_{i=1}^{n}$. But after feed-forward process in the
		feature space, the noise $\eta$ is reduced to very small scale
		and we got $A_n = \{a_i\}^n_{i=1} \approx \{ f(a_{i}) +
		f(\mu_{i})\}_{i=1}^{n}$. Therefore activation vectors are 
		concentrated around a noiseless local tangent hyperplane
		$\mathcal{P}$ of manifold $\mathcal{M}$.
		
		To realize the goal of estimating local tangent $d=dim\mathcal{P}$ of
		manifold $\mathcal{M}$ given $\tilde{X}_n$ and corresponding
		$A_{D,n}$ in a specific layer, we adopt the standard
		approach to compute the SVD, with the singular values (denoted as
		$\sigma$) yielding $j$. With high probability
		the first $j$ singular values are significant, so that
		$\sigma_{1}, \ldots, \sigma_{j}\gg\sigma_{j+1}, \ldots,
		\sigma_{D}$, we then take the reasonable estimation that $d=j$.

		\subsection{estimate the manifold dimension of different layers}

		\textbf{Fully connected layers.}
		For fully connected layers, the size of $A_{D \times n}$ would be
		$(D,n)$. Then we apply SVD on $A_{D \times n}$ to get its singular
		value array $\sigma$ that sorted in descending order. Let $\sigma_i$
		denotes the i-th singular value in the array, if there exist some
		$j > 0$, where $ \sigma_j/\sigma_{j+1} > \theta $, in which
		$\theta$ is a very big number, then we claim that $j$ is an
		estimation of the tangent hyperplane $\mathcal{P}$'s dimension,
		so is the estimation of the local dimension of manifold
		$\mathcal{M}$ of corresponding layer and original images. 
		If the $j$ does not exist, then the estimation of the local dimension
		would be up to the dimension of the whole activation space(see
		Section \ref{support_exp}). 
			
		\textbf{Convolutional layers.}
		We denote $D=H \times W \times C$ as the activation space
		dimension of a specific layer with whose $H$ is feature map's
		height, $W$ is the width of feature map , $C$ is the number of
		channel. For data cluster with size $n$ and the $i$th feature map
		with height $H$ and width $W$ in the convolutional layer we got a
		corresponding matrix $A_{(H \times W),i,n}$ and calculate
		dimension respectively.
		
\begin{figure}[!h]
			\begin{center}
			\vspace{-2em}
			\includegraphics[width=0.8\textwidth]{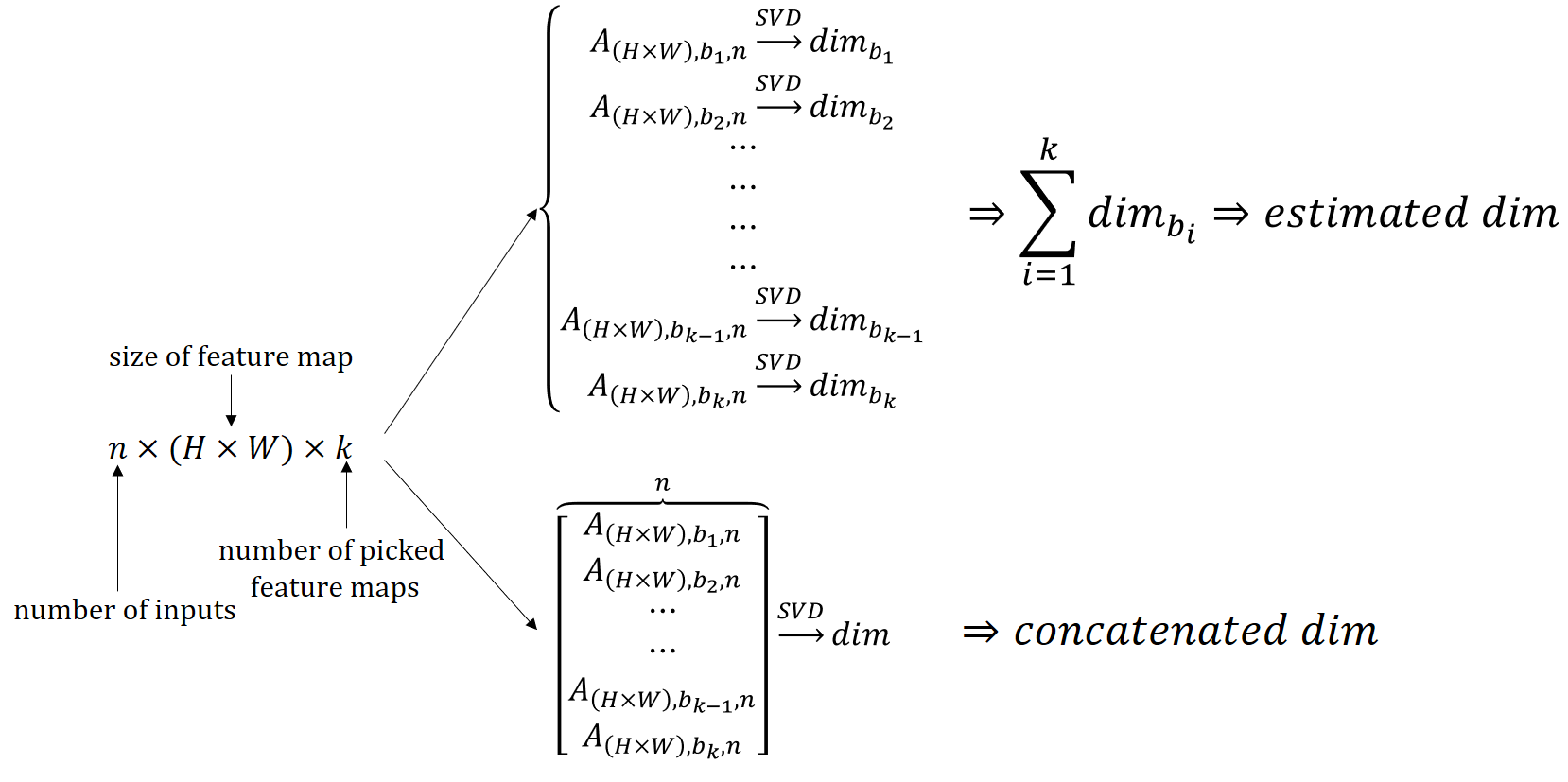}
			\caption{How to compute the estimated dimension and
				concatenated dimension. $b_i$ denotes the index of i-th
				randomly picked feature map. $1 \leq i \leq k$. The matrix
				in the bottom consists of all the $A_{(H \times W),b_i,n}$
				as submatrices.}
			\label{figsum_svd_vs_concatenated_svd} 
			\vspace{-2em}
			\end{center}
		\end{figure}

		We define the \textit{estimated dimension} by randomly picking $k (k \geq
		1)$ feature maps and calculating their dimension one by one and sum
		the dimensions up. 
		\textit{Concatenated dimension}:
		concatenate the $k$ picked feature maps and calculate the
		concatenated matrix's dimension.
		\textit{Original dimension}: 
		when we pick all the feature maps in a layer and
		calculate the concatenated dimension, this \textit{concatenated dimension} is defined as the \textit{original dimension}. 
		Figure \ref{figsum_svd_vs_concatenated_svd} is a illustration of these concepts.
		
		We should note that fully-connected layers' are 1 ($C=1$),
		so in fully-connected layer, the \textit{estimated dimension} =
		\textit{concatenated dimension} = \textit{original dimension}. We
		will refer to either of it for the same meaning. When we refer to
		a layer's \textit{estimated dimension} (or \textit{estimated
			dimension} of a layer), we mean we choose all feature maps and
		do calculation ($k=C$).

\section{Experimental Setup}
	\label{secexpsetting}

	\textbf{Network.} VGG19 was proposed by \citet{2014arXiv1409.1556S}, it was proved
	to have excellent performance in the ILSVRC competition, whose
	pre-trained model are still wildly used in many areas. In this
	paper, we use pre-trained VGG19 model to conduct our experiment,
	and give every layer a unique name so that they can be referred
	to directly. Table \ref{tblVGG19} in Appendix gives the name of every layer
	and its corresponding activation space dimension.

	\textbf{Image augmentation.} 
	We choose three categories in ImageNet  dataset: Persian Cat
	(n02123394), Container Ship (n03095699) and Volcano (n09472597) (See Figure \ref{figori_images} in Appendix). 
	Then for each category, we select three typical images with
	high probability as the original images, because the network
	firmly believe that they are in the category, their activation
	vectors can be considered as a data point on the same manifold
	representing Persian Cat.

\begin{figure}[!h]
	\begin{center}
		\begin{subfigure}[]{0.22\textwidth}
			\includegraphics[width=0.9in]{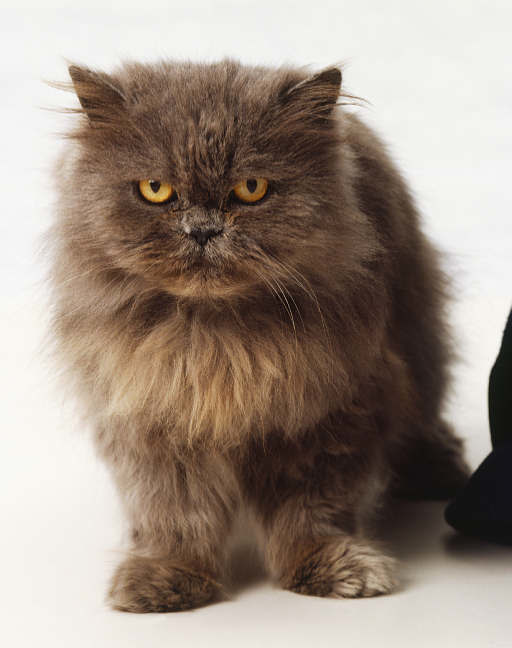}
			\caption{Original image.}
		\end{subfigure}	
		\begin{subfigure}[]{0.22\textwidth}
			\includegraphics[width=0.85in]{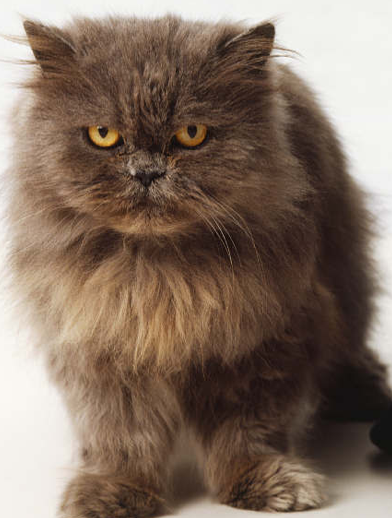}
			\caption{Cropping.}
		\end{subfigure}
		\begin{subfigure}[]{0.22\textwidth}
			\includegraphics[width=0.9in]{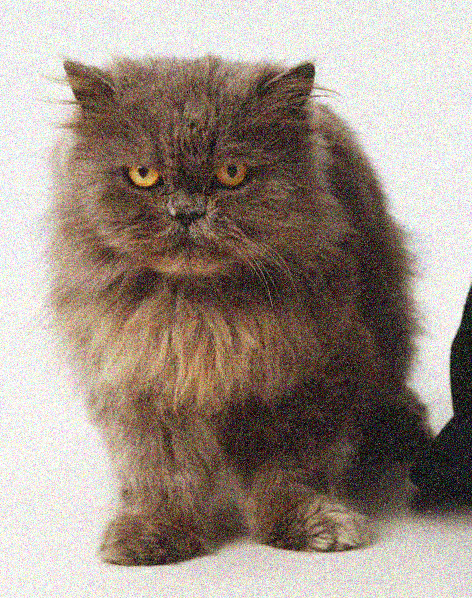}
			\caption{Gaussian noise.}
		\end{subfigure}
		\begin{subfigure}[]{0.22\textwidth}
			\includegraphics[width=0.9in]{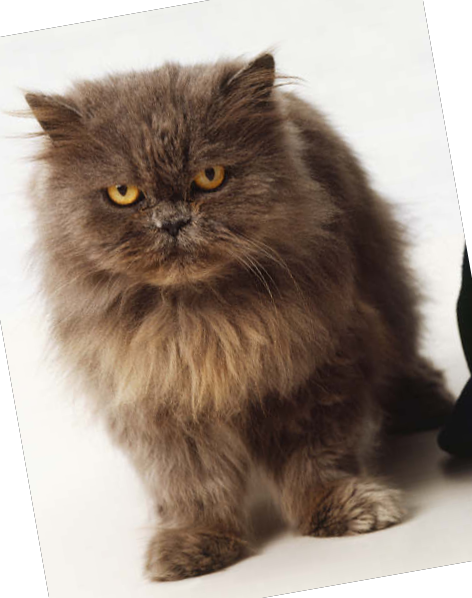}
			\caption{Rotation.}
		\end{subfigure}
		\caption{Cropping, Gaussian noise, rotation on an original image.
		We exaggerate the methods to show the change.}
		\label{figgen_methods} 
	\vspace{-2em}
	\end{center}
	\end{figure}

	We use three augmentation methods to generate similar images
	which form a data point cluster, they are:
	1) Cropping: randomly cut out a strip of few pixels from every edge;
	2) Gaussian noise: add Gaussian noise (mean =
	0.0, var = 0.01) to every single pixel; 
	3) Rotation: rotate the image by a random degree
	in $\pm 10\%$. The exaggerated output
	images of these three methods are shown in Figure
	\ref{figgen_methods}.

	As these augmentation methods only apply small changes on the
	original image $X$ (also keep high probability $P(x)>0.99$), activation vectors 
	$A$ will concentrate around the activation
	vector $a_0$ of $x_0$, which can be considered near a local
	tangent hyperplane $\mathcal{P}$ of the manifold $\mathcal{M}$.

\section{Supported Experiments}
	\label{secsupexp}
	\label{support_exp}

	We have tried three Persian cat images, the dimension is within a small range.
	For other categories, we also tried three high-probability images for ship and three high-probability images for volcano.  
	The dimension of vocano is slightly higher than that of cat, and the dimension of cat is slightly higher than ship for the same layer.
	All the three category show the same trend on dimension through the layers.
	Therefore, we will show the details mainly on a typical Persian cat as shown in Figure \ref{figgen_methods} (a). 
	
\begin{figure}[!h]
		\vspace{-0.5em}
		\begin{center}
		\begin{minipage}[!h]{0.49\textwidth}
			\begin{subfigure}[]{2.4in}
				\includegraphics[width=\textwidth]{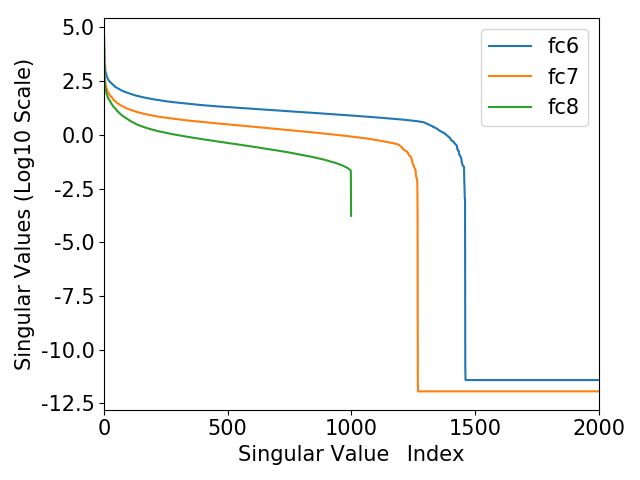}
				\caption{Cropping.}
				\vspace{\baselineskip}
			\end{subfigure}	

			\begin{subfigure}[]{2.4in}
				\includegraphics[width=\textwidth]{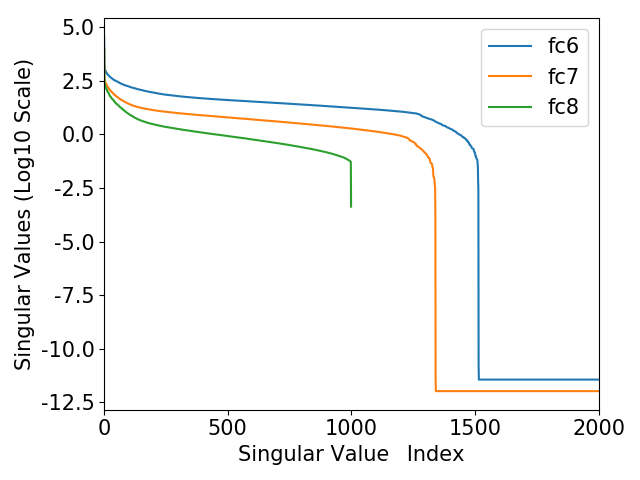}
				\caption{Gaussian noise.}
				\vspace{\baselineskip}
			\end{subfigure}	
			
			\begin{subfigure}[]{2.4in}
				\includegraphics[width=\textwidth]{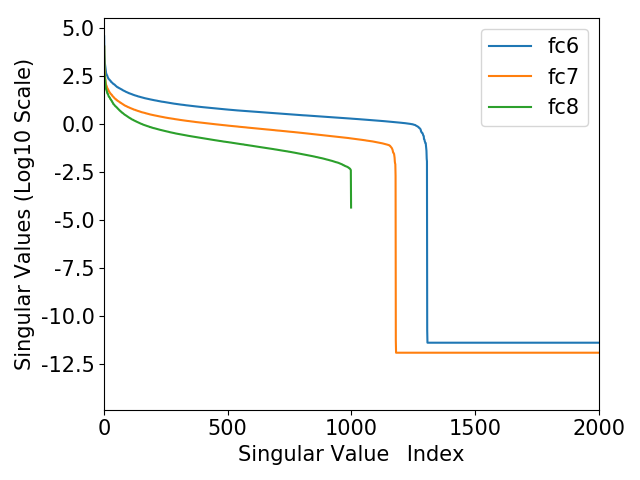}
				\caption{Rotation.}
				\vspace{\baselineskip}
			\end{subfigure}
		\caption{``Drops" for fully connected layers. \\
			fc6 and fc7 have dramatic drops.} 
		\label{figsingular_value_drop} 
		\end{minipage}
		\begin{minipage}[!h]{0.49\textwidth}
			\begin{subfigure}[]{2.4in}
					\includegraphics[width=\textwidth]{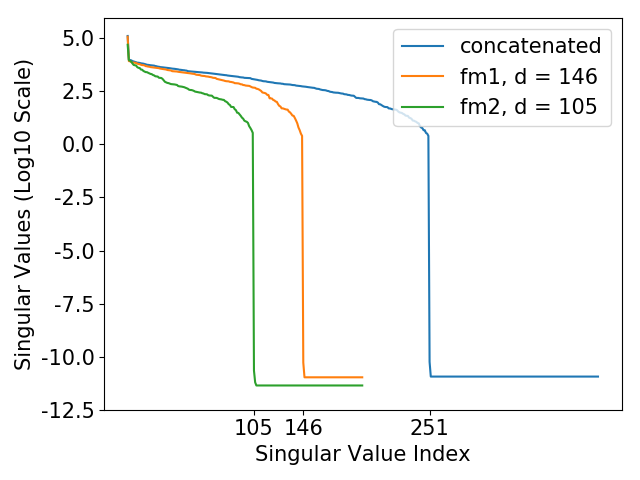}
					\caption{Estimated dimension $ = 105 + 146 = 251 $ \\
						= concatenated dimension.}
				\end{subfigure}	
				
				\begin{subfigure}[]{2.4in}
					\includegraphics[width=\textwidth]{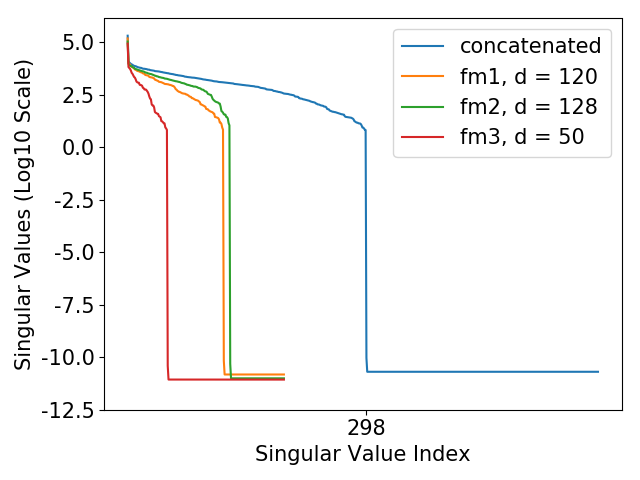}
					\caption{Estimated dimension $ = 120 + 128 + 50 = 298 $
					= concatenated dimension.}
				\end{subfigure}	
				
				\begin{subfigure}[]{2.4in}
					\includegraphics[width=\textwidth]{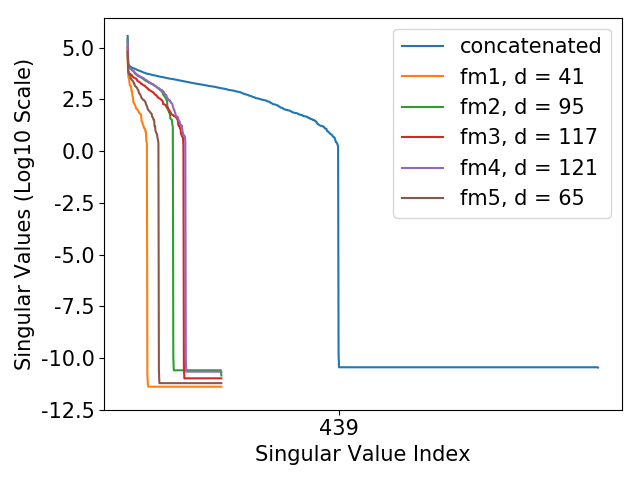}
					\caption{Estimated dimension $ = 41 + 95 + 117 + 121 + 65
					= 439 $ = concatenated dimension.}
				\end{subfigure}	
		\caption{Estimated dimension and concatenated dimension for Conv5\_1 , $k \in \{2, 3, 5\}$.}
			\vspace{\baselineskip} 
			\label{figfeatureMapConcat} 
			\end{minipage}
		\end{center}
	\vspace{-1em}
	\end{figure}

	\textbf{Estimated dimension for a fully connected layer or a single
		feature map.}
	We apply SVD on $A_{D \times n}$ in fully connected layers and then 
	plot the singular values in log10 scale (
	Figure \ref{figsingular_value_drop}). If we
	specify a certain layer, we can find dramatic drop at $j$ for
	all the three augmenting methods: $\sigma_j/\sigma_{j+1}>\theta=10^5$. So we
	can claim that for the local tangent hyperplane $\mathcal{P}$ on
	manifold $\mathcal{M}$, the dimension is $d=j$ with high probability
	as long as we use enough samples (See Section \ref{secSVD}). 
	This ``dramatic drop'' also appears for a single feature map. 
	The only exception is on the fc8 layer, there is no ``dramatic drop'',
	inferring that the hyperplane $\mathcal{P}$ spans
	the whole activation space. $d=1000$ in fc8.
	
	\textit{Rule 1}. The \textit{estimated dimension} of the local tangent
	hyperplane of the manifold for a fully connected layer or a single feature map can be determined by a dramatic drop along the singular values.

\begin{figure}[!h]
		\begin{center}
			\includegraphics[width=4in]{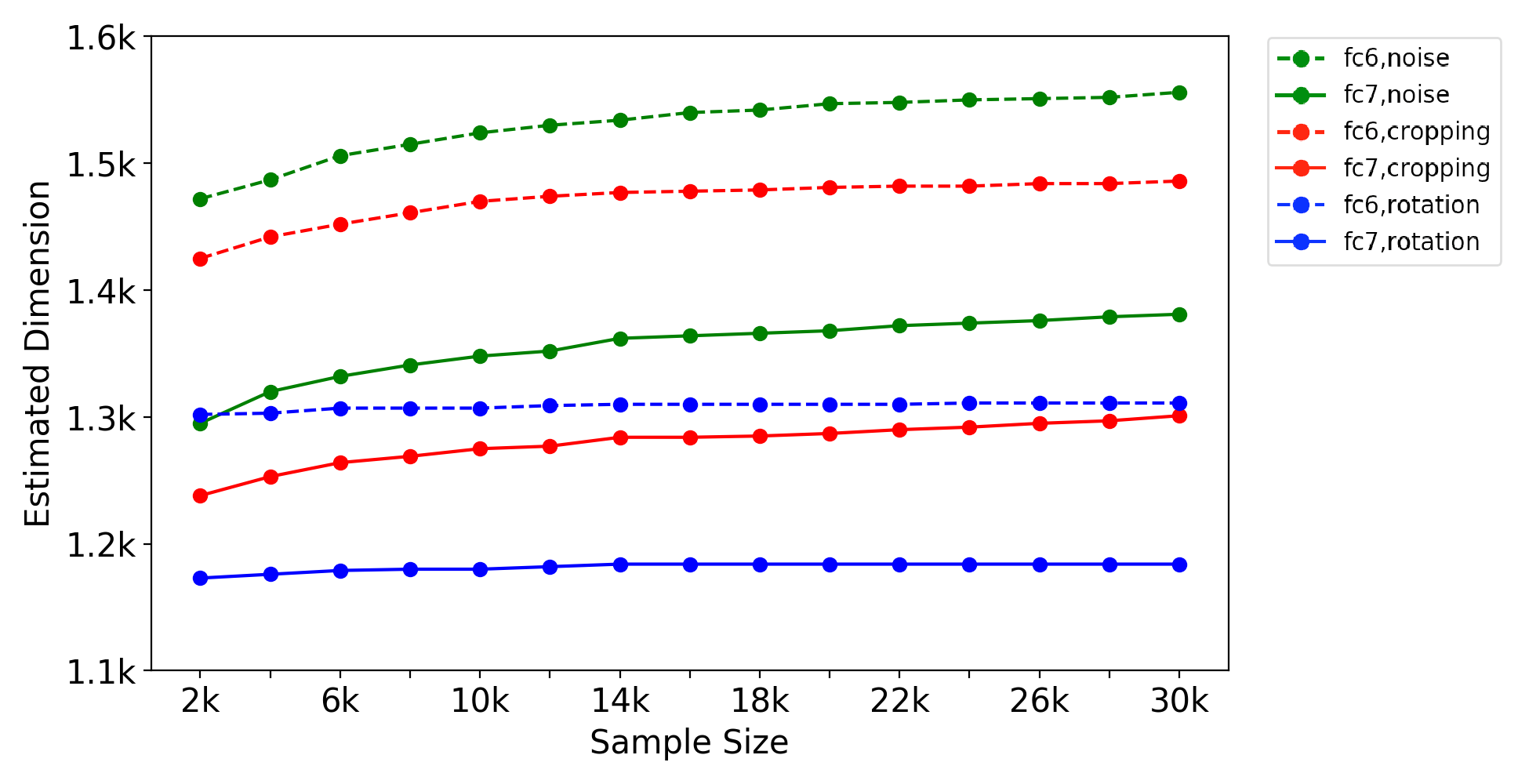}
			\caption{Estimated dimension versus data cluster size in fully connected layer.}
			\label{figdimen_size_fc} 
	
			\begin{minipage}[!h]{0.49\textwidth}				
			\includegraphics[width=2.5in]{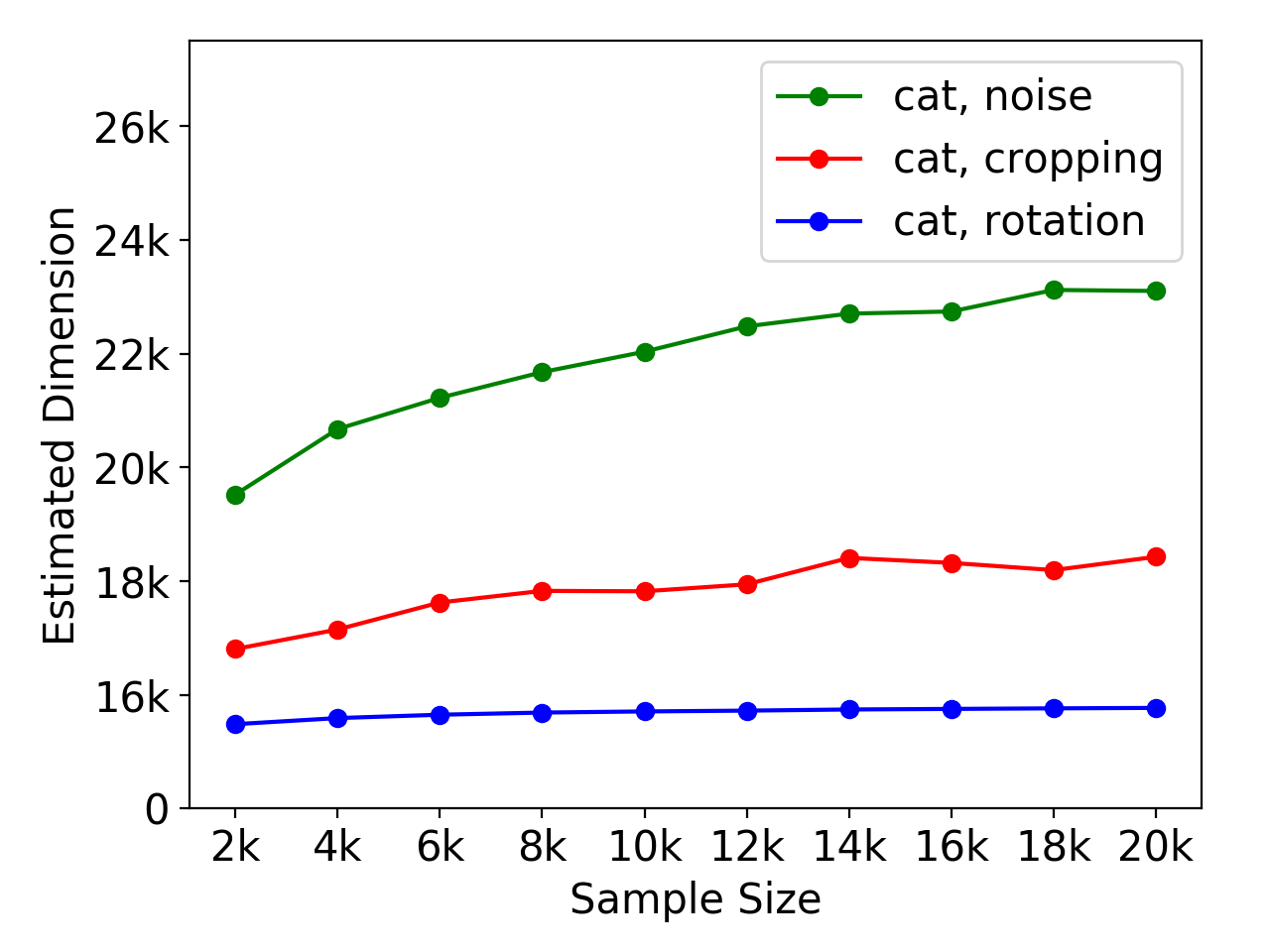}
			\caption{Estimated dimension versus data \\
				size in Conv5\_4 layer.}
			\label{figdimen_size_conv5_4} 
			\end{minipage}
			\begin{minipage}[!h]{0.49\textwidth}				
			\includegraphics[width=2.5in]{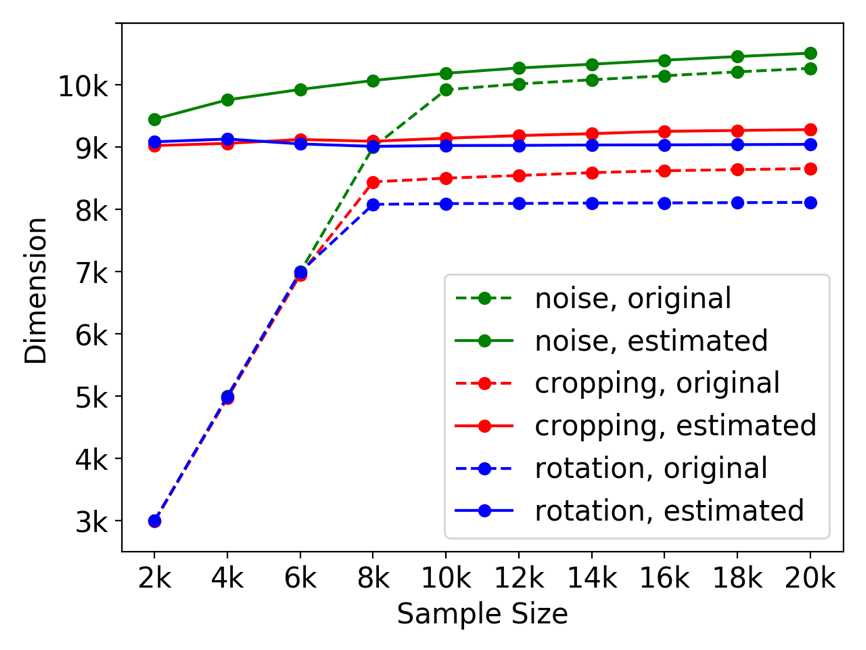}
			\vspace{-1em}
			\caption{Estimated dimension and original \\ 
			dimension for maxpooling5.}
			\label{figest_ori_dim} 
			\end{minipage}
		\end{center}
	\vspace{-1em}
	\end{figure}

	\textbf{Influence on the estimated dimension by the data size.}
	Figure \ref{figdimen_size_fc} shows the \textit{estimated dimension}
	versus the augmentation data size. The dimension grows slowly with the growth of the data size. Although the augmentation data scale
		influences the estimated dimension, the growth on the
	dimension along with the data size is fairly small. The dimension only
	grows by less than 3\% as the data size triples. 
	Thus, it is reasonable to use a fairly
	small data set to estimate the dimension. More importantly, as shown in Figure \ref{figdimen_size_conv5_4},
	such rule can also be generalized to calculate the \textit{estimated dimension} of the convolutional
	layers.  
		
	\textit{Rule 2.} We can determine the local tangent
	hyperplane's \textit{estimated dimension} of the manifold in a
	layer (fully connected or convolution) using a fairly small
	data cluster size, for example, 8k.
	
	\textbf{Estimated dimension and concatenated/original dimension.} 
	We randomly pick $k \in \{2, 3, 5, 10\}$ feature maps in layer Conv5\_1 and calculate the \textit{estimated dimension}
	as well as \textit{concatenated dimension}. The result of $ k \in \{2, 3,
		5\}$ are as shown in Figure \ref{figfeatureMapConcat}.
	For the result of $ k = 10 $, see Table \ref{tblfeature_map_concat} in Appendix. 
		The \textit{estimated
		dimension} is very close to the corresponding \textit{concatenated
		dimension}. Thus, we can use the \textit{estimated dimension} to
	approximate the \textit{concatenated dimension}.
	
	We then pick all feature maps in
	maxpooling5, and calculate the \textit{estimated dimension, original dimension}. 
	Figure \ref{figest_ori_dim} shows
	that start from 8k of the data size, the \textit{estimated
		dimension} is close to the \textit{original dimension}. 
	Thus, we can use a small amount of 8000 images to approximate the \textit{original
		dimension} using the \textit{estimated dimension}.
	
	When the data cluster size is insufficient, assuming the local
	tangent hyperplane of the manifold is $d$-dimensional, the result
	will be strictly restricted by the number of input images $n$
	when $n < d$. So that the \textit{concatenated dimension} or
	\textit{original dimension} we calculate would be almost equal to
	$n$ for small $n$, while \textit{estimated dimension} is a
	summation which can approximate $d$.
	
	\textit{Rule 3.} The \textit{original dimension} of the local tangent
	hyperplane can be approximated by the \textit{estimated dimension} using a fairly small size of dataset,
	for example 8000. 
	
	\section{Dimensions of the Deep Manifolds}
	\label{secfinal_result}
	For each of the three categories, Persian Cat (n02123394), Container Ship (n03095699) and Volcano (n09472597) in ImageNet, we randomly choose three images of high probability and determine the estimated
	dimensions based on the three rules drawn in Section \ref{support_exp}. 
	
\begin{figure}[!h]
			\vspace{-1em}
			\begin{center}
				\begin{minipage}[!h]{0.49\textwidth}
					\begin{center}
					\includegraphics[width=2in]{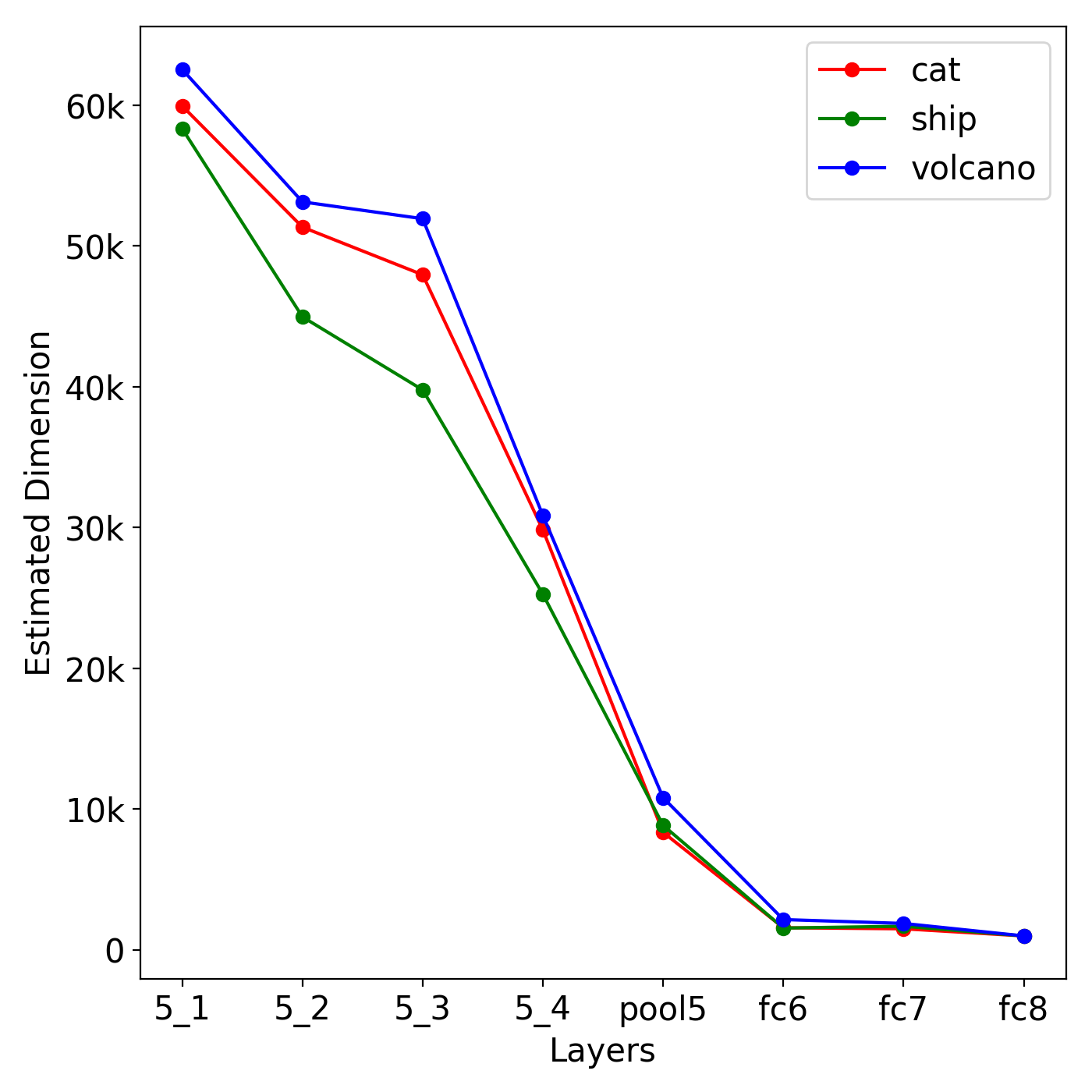}
					\caption{Dimension of Conv5 and fc layers.}
					\label{figtrend_conv5_fc} 
					\end{center}
				\end{minipage}
				\begin{minipage}[!h]{0.49\textwidth}
					\begin{center}
					\includegraphics[width=2in]{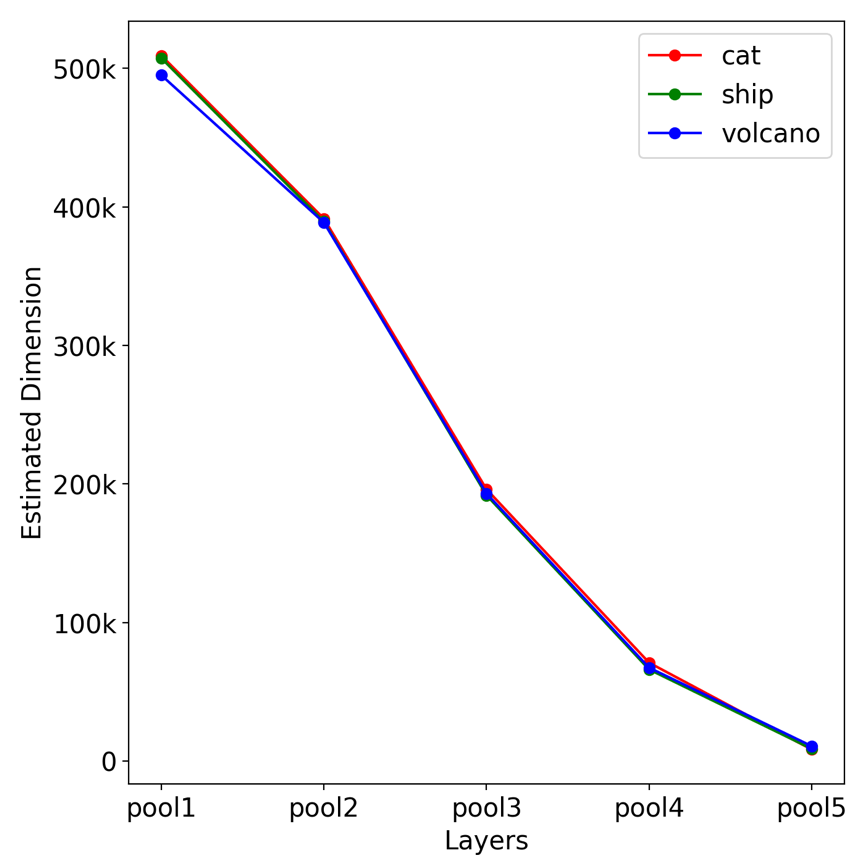}
					\caption{Dimension of all maxpooling layers.}
					\label{figtrend_maxpool} 
					\end{center}	
				\end{minipage}
			\end{center}
			\vspace{-1em}						
		\end{figure}

	\textbf{Dimensions for Conv5 and fully connected layers}. For Conv5 and fully connected layers, we summarize the average of the estimated
	dimensions in Table \ref{tblrst_conv5_fc} and Figure \ref{figtrend_conv5_fc}. 
	The \textit{estimated dimension}
	gradually declines from Conv5\_1 to fc8. 
	For fc6 and fc7, the activations lie in a low-dimension manifold
		embedded in the 4096-dimension space. For fc8, the manifold's dimension is exactly 1000. 
		It makes sense as fc8 is directly linked to the final classification prediction, it is in full rank to achieve a higher performance.
		The dimensions of the three categories are close to each other and decline quickly inside the four convolutional layers and the last maxpooling layer. 
		
		\begin{table}[!ht]
		\begin{center}
			\resizebox{0.8\textwidth}{!}{
				\begin{minipage}{\textwidth}
				\begin{center}
			\caption{Dimensions of Conv5 and fully connected layers.}
			\label{tblrst_conv5_fc}
			\begin{tabular}{|c|r|r|r|r|r|r|r|r|}
				\hline
						& \multicolumn{1}{c|}{5\_1} & \multicolumn{1}{c|}{5\_2} & \multicolumn{1}{c|}{5\_3} & \multicolumn{1}{c|}{5\_4} & \multicolumn{1}{c|}{pool5} & \multicolumn{1}{c|}{fc6} & \multicolumn{1}{c|}{fc7} & \multicolumn{1}{c|}{fc8} \\ \hline
				cat     & 59946                     & 51347                     & 47958                     & 29834                     & 8358                       & 1580                     & 1506                     & 1000                     \\ \hline 
				ship    & 58329                     & 44968                     & 39781                     & 25267                     & 8851                       & 1577                     & 1691                     & 1000                     \\ \hline 
				volcano & 62540                     & 53136                     & 51939                     & 30862                     & 10816                      & 2163                     & 1887                     & 1000                     \\ \hline 
			\end{tabular}
			\end{center}
		\end{minipage}
			}
		\end{center}
		\end{table}
	
	\textbf{Dimensions for maxpooling layers.} We illustrate the average of the estimated
	dimensions in Figure \ref{figtrend_maxpool} all maxpooling layers. The dimensions of the three categories coincide with each other and decline quickly for deep pooling layers.
	
	

	\section{Conclusions}
	\label{secresult}


	Through extensive experiments, we found that there exists a dramatic drop for the singular values of the fully connected layers or a single feature map of the convolutional layer, and the dimension of the concatenated feature vector almost equals the summation of the dimension of each feature map for several feature maps randomly picked. 
	Based on the interesting observations we obtained, we developed an efficient and effective SVD based method to estimate the local dimension of deep manifolds in the VGG19 neural network.  
	We found that the dimensions are close for different images of the same category and even images of different categories, and the dimension declines quickly along the convolutional layers and fully connected layers.
	Our results support the low-dimensional manifold hypothesis for deep networks, and our exploration helps unveiling the inner organization of deep networks.
	Our work will also inspire further possibility of observing every feature map separately for the dimension of convolutional layers, 
	rather than directly working on the whole activation feature maps, which is costly or even impossible for the current normal computing power.
	
	\subsubsection*{Acknowledgments}
	The work is supported by National Natural Science Foundation of China (61772219, 61472147) and US Army Research Office (W911NF-14-1-0477). 
	
	\bibliography{iclr2018_conference}
	\bibliographystyle{iclr2018_conference}
	
	\newpage
	
	\section*{Appendix}
	\label{secappendix}
	
	\begin{figure}[!h]
		\begin{center}
			\includegraphics[width=1.1in]{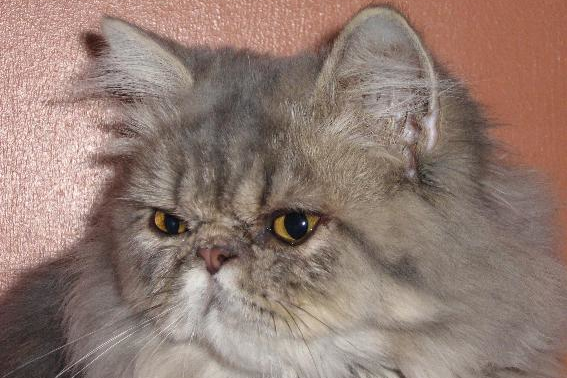}
			\includegraphics[width=1.1in]{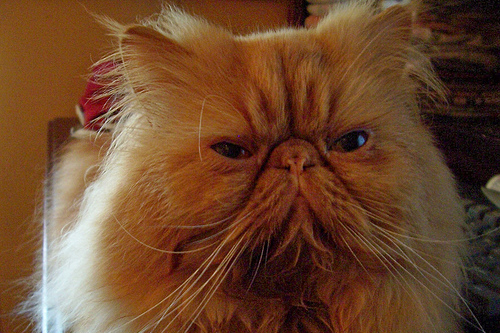}
			\includegraphics[width=1.1in]{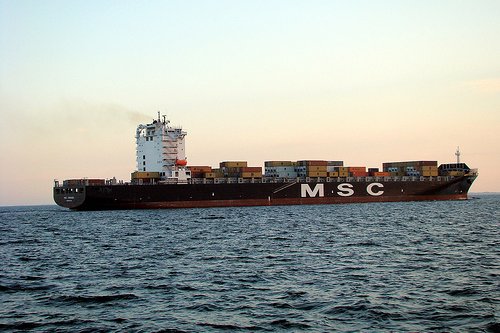}
			\includegraphics[width=1.1in]{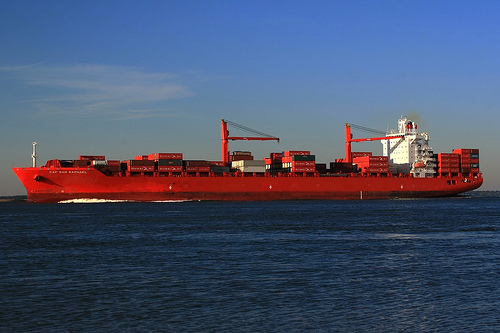}
			
			\includegraphics[width=1.1in]{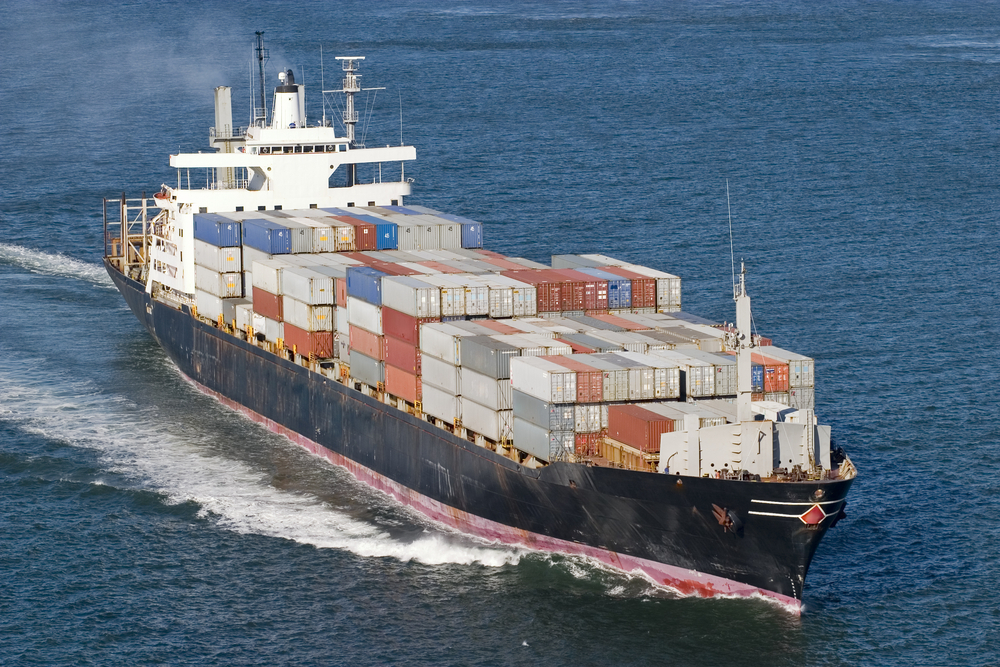}
			\includegraphics[width=1.1in]{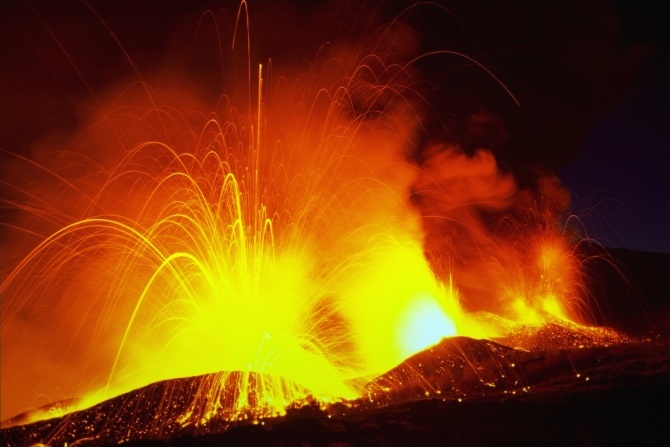}
			\includegraphics[width=1.1in]{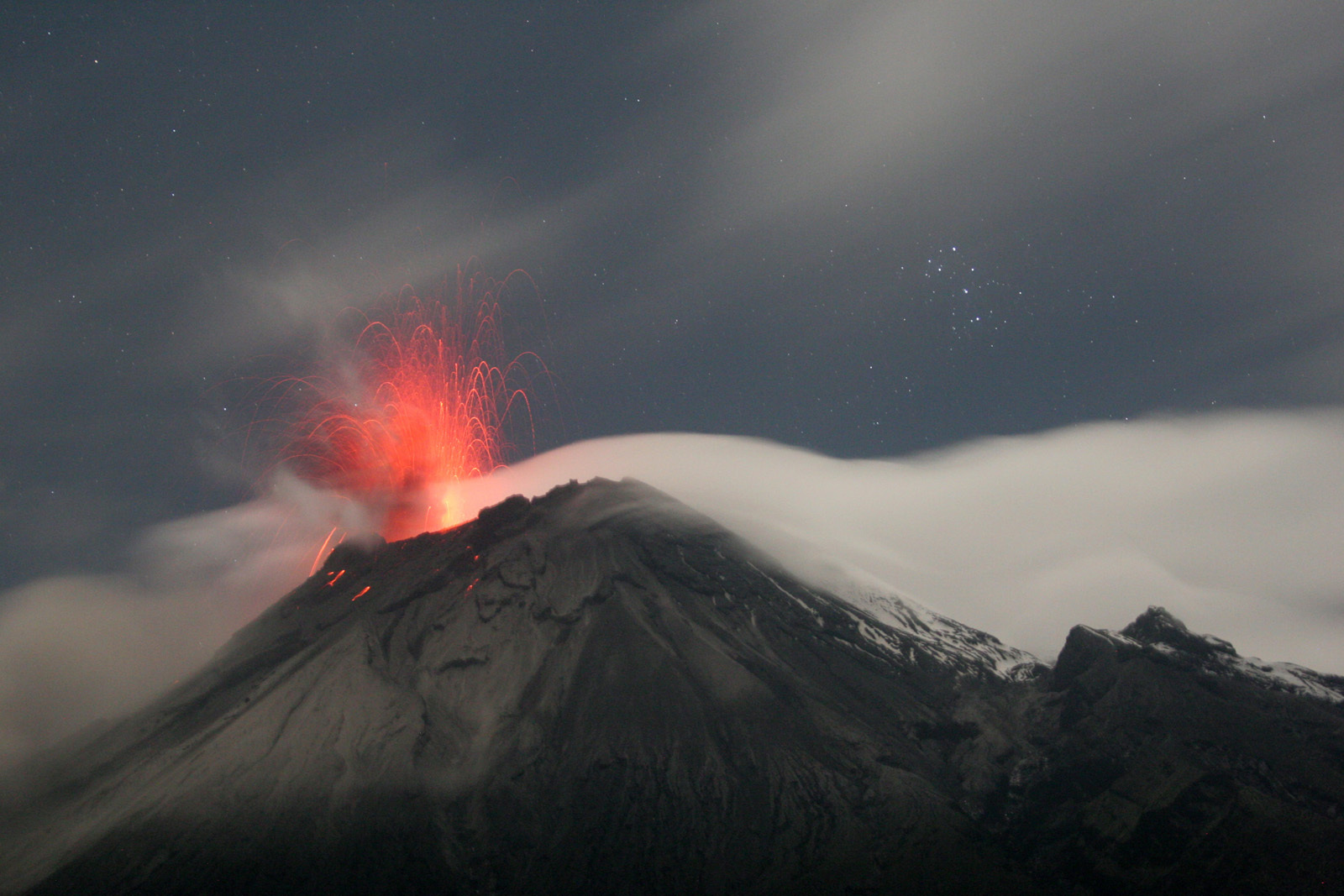}
			\includegraphics[width=1.1in]{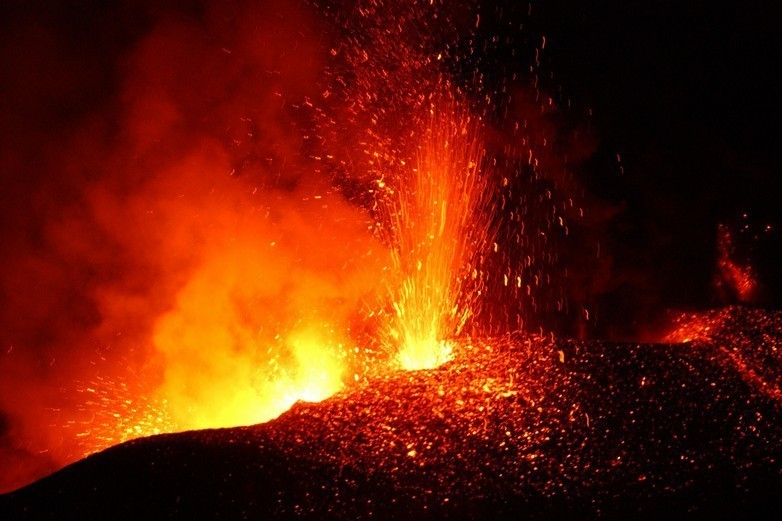}
			\caption{More original images for cat, ship and volcano. Cropped to make them aligned. } 
			\label{figori_images} 
		\end{center}
	\end{figure}

	\begin{table}[!ht]
		\begin{center}
			\caption{VGG19 network architecture (dimension of activation space = size of feature maps $\times$ number of channels).}
			\label{tblVGG19}
			\begin{tabular}{|c|r|r|r|}
			\hline
			\multicolumn{1}{|c|}{Layer}  	 & \multicolumn{1}{c|}{\begin{tabular}[c]{@{}c@{}}Size of \\ feature maps\end{tabular}} & \multicolumn{1}{c|}{\begin{tabular}[c]{@{}c@{}}Number of\\   channels\end{tabular}} & \multicolumn{1}{c|}{\begin{tabular}[c]{@{}c@{}}Dimension of\\   activation space\end{tabular}} \\ \hline \hline
			conv1 $(1\_1, 1\_2)$           	 & 224$\times$224                                                                       & 64                                                                                  & 3211264                                                                                        \\ \hline
			maxpooling1                    	 & 112$\times$112                                                                       & 64                                                                                  & 802816                                                                                         \\ \hline
			conv2 $(2\_1, 2\_2)$           	 & 112$\times$112                                                                       & 128                                                                                 & 1605632                                                                                        \\ \hline
			maxpooling2                    	 & 56$\times$56                                                                         & 128                                                                                 & 401408                                                                                         \\ \hline
			conv3 $(3\_1, 3\_2, 3\_3, 3\_4)$ & 56$\times$56                                                                         & 256                                                                                 & 802816                                                                                         \\ \hline
			maxpooling3                    	 & 28$\times$28                                                                         & 256                                                                                 & 200704                                                                                         \\ \hline
			conv4 $(4\_1, 4\_2, 4\_3, 4\_4)$ & 28$\times$28                                                                         & 512                                                                                 & 401408                                                                                         \\ \hline
			maxpooling4                    	 & 14$\times$14                                                                         & 512                                                                                 & 100352                                                                                         \\ \hline
			conv5 $(5\_1, 5\_2, 5\_3, 5\_4)$ & 14$\times$14                                                                         & 512                                                                                 & 100352                                                                                         \\ \hline
			maxpooling5                    	 & 7$\times$7                                                                           & 512                                                                                 & 25088                                                                                          \\ \hline
			fc6                          	 & 4096                                                                                 & 1                                                                                   & 4096                                                                                           \\ \hline
			fc7                          	 & 4096                                                                                 & 1                                                                                   & 4096                                                                                           \\ \hline
			fc8                          	 & 1000                                                                                 & 1                                                                                   & 1000                                                                                           \\ \hline
			\end{tabular}
		\end{center}
		\end{table}

		\begin{table}[!h]
			\begin{center}
			\caption{Estimated dimension and concatenated dimension in Conv5$\_1$ layer for $k = 10$. 
			Dim.$i$ ($i \in {0,1,2,...,9}$) is the estimated dimension of the $i$-th randomly picked feature maps.
			Dim.estim = $\sum_{i=0}^9$ Dim.$i$ 
			and Dim.concat is the concatenated dimension of the $k$ feature maps.
			} 
			\label{tblfeature_map_concat}
			\begin{tabular}{|l|r|r|r|}
				\hline
				\multicolumn{1}{|c|}{} & \multicolumn{1}{c|}{Sample1} & \multicolumn{1}{c|}{Sample2} & \multicolumn{1}{c|}{Sample3} \\ \hline
				Dim.0                  & 124                          & 131                          & 158                          \\ \hline
				Dim.1                  & 112                          & 163                          & 80                           \\ \hline
				Dim.2                  & 141                          & 126                          & 62                           \\ \hline
				Dim.3                  & 128                          & 95                           & 120                          \\ \hline
				Dim.4                  & 141                          & 156                          & 61                           \\ \hline
				Dim.5                  & 78                           & 129                          & 155                          \\ \hline
				Dim.6                  & 96                           & 152                          & 112                          \\ \hline
				Dim.7                  & 113                          & 103                          & 116                          \\ \hline
				Dim.8                  & 123                          & 62                           & 118                          \\ \hline
				Dim.9                  & 157                          & 120                          & 128                          \\ \hline
				Dim.estim			   & 1213 						  & 1237						 & 1110							\\ \hline
				Dim.concat             & 1213                         & 1237                         & 1110                         \\ \hline
			\end{tabular}
	\end{center}
	\end{table}
	
\end{document}